\documentclass{bmvc2k}

\title{TraMP-LLaMA: Generative Interpretability with Decoupled Instruction Tuning for Facial Expression Quality Assessment}

\addauthor{Shuchao Duan}{shuchao.duan@bristol.ac.uk}{1}
\addauthor{Alan Whone}{alan.whone@bristol.ac.uk}{2}
\addauthor{Hossein Rahmani}{h.rahmani@lancaster.ac.uk}{3}
\addauthor{Jun Liu}{j.liu81@lancaster.ac.uk}{3}
\addauthor{Majid Mirmehdi}{m.mirmehdi@bristol.ac.uk}{1}

\addinstitution{
 School of Computer Science\\
 University of Bristol\\
 Bristol, UK
}
\addinstitution{
 Translational Health Sciences\\
 University of Bristol\\
 Bristol, UK
}
\addinstitution{
  School of Computing and Communications\\
 Lancaster University\\
 Lancaster, UK
}

\runninghead{Duan et al.}{ Generative Interpretability for FEQA}

\usepackage[T1]{fontenc}
\usepackage[utf8]{inputenc} 
\usepackage{textcomp}

\usepackage{dblfloatfix}
\usepackage{amssymb}
\usepackage{xcolor}
\usepackage{array}

\usepackage{amsmath}

\usepackage{booktabs}
\usepackage{placeins}

\usepackage{graphicx}  
\usepackage{rotating}  %

\usepackage{booktabs}
\usepackage{multirow}
\usepackage{ulem} 

\begin{document}

\maketitle

\begin{abstract}
Existing facial expression quality assessment (FEQA) methods typically produce only a severity score, without explicitly communicating the observable facial motion evidence that supports the prediction. This limits interpretability and makes it difficult to inspect the basis of model outputs in Parkinson’s disease assessment. To address this gap, 
we propose TraMP-LLaMA, a unified multimodal framework that jointly predicts severity scores and generates structured textual reports from facial motion cues. The framework integrates RGB appearance and landmark trajectory cues, and adopts a decoupled instruction-tuning strategy to reduce task interference between severity prediction and language generation. To support this task, we further extend the PFED5 dataset with expert-guided textual motion descriptions and construct PFED5+. Experiments on PFED5+ show that TraMP-LLaMA outperforms competitive video-language baselines in report generation and achieves the best severity prediction performance among the compared methods under joint multi-expression training, improving Spearman’s rank correlation by at least 4.39\% over all competing methods. The text annotations and code are available at \url{https://github.com/shuchaoduan/TraMP-LLaMA}.

\end{abstract}

\section{Introduction}
\label{sec:intro}

\begin{figure}[t]
\centering
    \includegraphics[width=\linewidth]{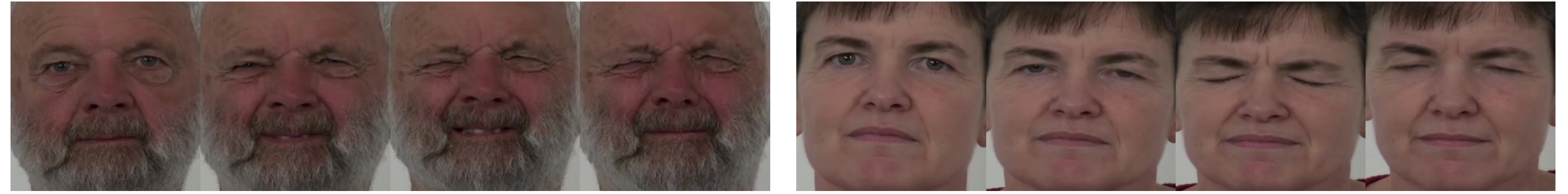}
    \caption{{\bf Same severity level, different evidence patterns for `Squeeze eyes'.} Both cases are assigned the same rating (level 2), but are supported by different facial motor evidence. The left case shows strong eyelid tightening with lips parted, whereas the right case exhibits milder tightening with minimal lower-face involvement.}
    \label{fig:samescore}
    \vspace{-3mm}
\end{figure}

Video-based facial expression quality assessment (FEQA) has emerged as a promising approach  for quantifying subtle facial motor impairments associated with neurological disorders, such as Parkinson’s disease (PD) \cite{duan2023qafe} and amyotrophic lateral sclerosis (ALS) \cite{torontoNeuroface}, as well as other clinically relevant affective states, e.g. pain \cite{pain_dataset_2011}.
However, despite recent progress, most existing FEQA approaches \cite{zhou2018visually,de2020deep,liu2023net,Szczapa_2022_trajec,moshkova2022assessment,ipapo2023clinical,duan2023qafe,liu2024hierarchical,tramp-former} ultimately return only a single severity score. While such outputs support quantitative 
monitoring, a score alone can only be interpreted through predefined clinical descriptors, such as MDS-UPDRS~\cite{mds-updrs2008}, which are coarse and generic and do not preserve the specific evidence needed for later audit or review. As shown in Fig.~\ref{fig:samescore} for `Squeeze eyes', two clips may receive the same rating, level 2, despite different facial regions and movement patterns.



{Recording the observations behind a score, even in a short text form, can therefore make FEQA outputs more transparent and auditable. It links ratings to visible motor evidence, allows later review or rescoring, supports more informative longitudinal comparison, and helps analyse disagreements between raters or between human and model assessments \cite{Nasarian2023DesigningIM}. }


Traditional interpretability methods, such as saliency maps and other post-hoc spatial attribution techniques \cite{Selvaraju2016GradCAMVE, Ribeiro2016WhySI}, remain limited in this setting. Although they can indicate which facial region influenced a prediction, they do not describe the spatio-temporal motor patterns that matter clinically, such as reduced range of motion, persistent blinking, or subtle asymmetry. 
Conversely, off-the-shelf {Large Language Models (LLMs)} and {Video-Language Models (VLMs)} {can generate fluent explanations, but} may hallucinate clinically irrelevant or visually unsupported details when asked to justify a prediction \cite{Asgari2025AFT}. 


{In this paper, we address this limitation by proposing TraMP-LLaMA, a unified multi-modal framework that jointly predicts a severity score and generates a structured textual report. We deploy VideoLLaMA3~\cite{zhang2025videollama} as the base VLM to provide general vision-language capability, and extend it with 
{a landmark trajectories stream to encode facial motion features.} 
For severity prediction, the scoring head operates solely on the joint trajectory-RGB representation obtained by fusing trajectory and RGB features. For report generation, in contrast, the language model is instruction-tuned on the concatenation of evidence tokens derived from RGB appearance, trajectory dynamics, and their fused representation. 
A key technical challenge in this joint formulation is avoiding task interference. Jointly optimising language generation and severity regression can distort spatio-temporal feature learning and degrade assessment performance. We therefore introduce a decoupled instruction tuning strategy: both tasks condition on the same shared spatio-temporal evidence, but only the scoring objective is allowed to update the motion encoder and the fusion module, while gradients from the language objective are blocked. This preserves reliable scoring while still enabling clinically grounded report generation.


{The recently released PFED5~\cite{duan2023qafe} is a unique facial expression assessment dataset that  provides MDS-UPDRS scores, but lacks textual annotations of the evidence that underlines the score in each case. We construct} PFED5+ by augmenting PFED5 with expert-guided, clinically reviewed, clip-level motion descriptions. These annotations describe visible, region-focused motor patterns without directly encoding the severity label, enabling report supervision while reducing label leakage.}

Our main contributions can be summarised as: 
(i) We propose TraMP-LLaMA, a multimodal framework that jointly predicts severity scores and generates structured evidence reports, producing inspectable facial motion evidence alongside quantitative assessments;
(ii) We develop a decoupled instruction-tuning strategy that mitigates task interference between severity regression and report generation by blocking language-generation gradients from updating the motion encoder and fusion module, while allowing both tasks to condition on the same spatio-temporal evidence;
(iii) Experiments on PFED5+, {our annotation-augmented extension of PFED5,} demonstrate that TraMP-LLaMA achieves state-of-the-art severity prediction under joint multi-expression training, 
outperforming prior methods by at least 4.39\%, {while surpassing VLM baselines in report generation by various metrics.}

\section{Related works}
\label{sec:literature}
We briefly review recent literature relevant to our work, focusing on facial expression analysis from videos in healthcare, as well as VLMs for generative interpretability.

\textbf{Video-based Facial Expression Analysis in Healthcare --}
Numerous studies have explored video-based facial analysis for early disease detection, aiming to distinguish patients from healthy controls in disorders such as PD \cite{Bandini2017,rajnoha2018towards,jin2020diagnosing,razzouki2024early} and ALS \cite{oliveira2024video}. For example, Razzouki et al. \cite{razzouki2024early} identified hypomimia in early-stage PD by extracting optical flow and RGB facial features with a video  transformer and classifying patients using a Random Forest model. While such methods are useful for screening, they remain limited to coarse patient-control discrimination. Others have developed classification-based approaches that predict multiple intensity levels to evaluate clinically relevant affective states, such as pain \cite{thiam2020two,wu2024global} and depression \cite{hu2023detecting,mahayossanunt2023explainable,li2024tsffm}. For example, Thiam et al. \cite{thiam2020two} proposed a two-stream CNN-BiLSTM combining optical flow and motion history to capture spatio-temporal pain-related features for classifying pain levels. While more fine-grained than binary detection, categorical models still discretise impairment into bins, overlooking subtle gradations \cite{duan2023qafe}.

Continuous severity assessment, often referred to as FEQA, has therefore emerged as a clinically relevant direction, where models regress facial dynamics to scores on validated clinical scales \cite{zhou2018visually,de2020deep,liu2023net,distanceordering2021,xu2020AU,Szczapa_2022_trajec,liu2024hierarchical,liao2024sequence,grammatikopoulou2019detecting,moshkova2022assessment,ipapo2023clinical,de2024facial,duan2023qafe,tramp-former}. For example, Duan et al. \cite{tramp-former} combined facial trajectories with RGB features to improve regression of subtle motion variations across severity levels. However, despite differences in task formulation, existing methods across detection, multi-classification, and regression generally return only an outcome label, category, or severity score, without explicitly describing the observable spatio-temporal motor evidence supporting that output. This limits their usefulness for clinically interpretable assessment.

\textbf{Vision-Language Models for Video and Facial Behavior Description --}
Recent VLMs have made it increasingly feasible to translate videos into natural language, supporting tasks such as captioning, question answering, and open-ended reasoning \cite{Zhang2023VideoLLaMAAI,Maaz2023VideoChatGPTTD,li2024mvbench,Shen2024LongVUSA,zhang2025videollama}. For example, LongVU~\cite{Shen2024LongVUSA} improves long-video understanding through spatio-temporal adaptive compression, reducing token redundancy while preserving informative visual content under limited context length. Likewise, VideoChat2~\cite{li2024mvbench} improves video reasoning through staged alignment and instruction tuning on diverse video-language tasks.
While such models demonstrate strong general video-to-text capability, they are primarily trained on web-scale corpora dominated by coarse events and scene-level semantics, and therefore are not designed to describe fine-grained, low-amplitude, and temporally nuanced evidence in a faithful manner \cite{Asgari2025AFT,Hong2025MotionBenchBA}. This limitation is particularly problematic in our setting, where the relevant evidence lies in subtle facial motor patterns rather than broad scene content.

Others have moved facial expression analysis beyond categorical recognition towards language-based description \cite{Nezami2019ImageCU,Yuan2023DescribeYF,Li2024FacialAB,Xie2024EmoVITRE,Chaubey2025FaceLLaVAFE}. For example, EmoVIT \cite{Xie2024EmoVITRE} adapts an instruction-following VLM to affective reasoning and generates textual justifications for emotional states using Action Unit (AU)-to-language supervision. Similarly, Face-LLaVA \cite{Chaubey2025FaceLLaVAFE} introduces facial priors and region-guided attention to improve anatomically grounded facial descriptions.
While these approaches improve interpretability compared with label-only prediction, most are developed for affective analysis or general face understanding rather than clinical motor assessment. Their outputs are often image-centric, AU-centric, or optimised for emotion understanding, and may therefore under-specify the spatio-temporal motor signs required in neurological evaluation, such as reduced movement amplitude, delayed initiation, asymmetry, or abnormal coordination across facial regions. Moreover, although datasets such as MAFW provide textual descriptions, many existing methods \cite{liu_mafw_2022,Chen2024FineCLIPERMF,foteinopoulou_emoclip_2024,Zhao2024EnhancingZF} use text primarily as auxiliary supervision or an additional modality to improve recognition performance rather than as a target output for evidence-grounded reporting.

This direction is conceptually related to radiology report generation \cite{huang2023kiut,Sloan2024AutomatedRR,zhang2024generalist,Wang2025DiagnosticCB}, where models translate visual findings into clinically usable text rather than returning only a prediction. However, unlike radiology, the relevant evidence in our setting is not static anatomical structure but subtle facial motion unfolding over time. Existing video-language and facial description models therefore do not directly address the problem studied here, which is to generate evidence-grounded reports that remain tightly aligned with the facial motor patterns supporting a clinical severity rating.

\section{Language Annotation Pipeline}
\label{sec6:text_template}

\begin{figure}[htbp!]
    \centering
    \includegraphics[width=0.9\linewidth]{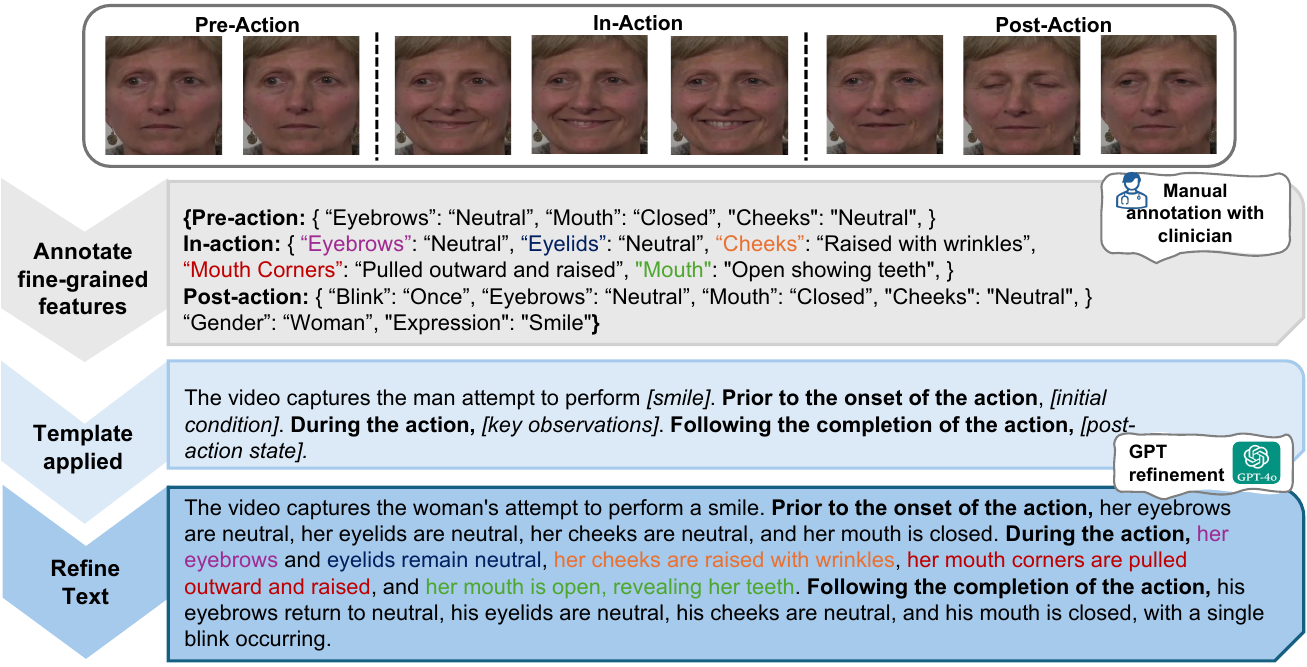}
    \caption{{\bf The pipeline for our structured annotation framework.} A video is first temporally segmented into \textit{pre-action}, \textit{in-action}, and \textit{post-action} phases. For each phase, clinically-relevant facial regions (e.g., eyebrows, mouth) are described according to a standardised template, generating a structured, factual description. The shown example is for the `Smile’ task from PFED5~\cite{duan2023qafe}.}
    \label{fig:text_generate_pipeline}
\end{figure}

To enable report generation grounded in observable motion evidence, we develop a systematic annotation framework that produces structured, clinically guided descriptions for each PFED5 video clip (see Fig.~\ref{fig:text_generate_pipeline}). {We call this extension PFED5+.} 
These new textual annotations introduced here explicitly describe observable motor evidence {(approved by two expert PD clinicians)} and are used as supervision for report generation.

\textbf{Temporal Structure and Standardised Template --} Facial-expression actions unfold over time and typically exhibit distinct phases, with phase boundaries following a consistent rule at the clip level. We therefore organise each annotation into three temporal components: 
(i) \textit{Pre-action} corresponds to the neutral baseline before visible motion onset; (ii) \textit{In-action} covers the main execution phase of the target movement; (iii) \textit{Post-action} describes the recovery stage after the movement peak, covering the offset and stabilisation of the face as it relaxes back toward a neutral state. Importantly, we do not annotate exact frame indices or timestamps; instead, phases are described at the clip level using this standard rule to ensure consistent coverage of the full temporal evolution. 
This three-phase template is enforced for all action clips, where each phase is described using the same set of region-specific slots. A complete template example is illustrated in Fig.~\ref{fig:text_generate_pipeline}.
When information in a phase is not observable due to incomplete face visibility, such as partial face crops or missing regions after preprocessing, we use a fixed placeholder (e.g., \textit{[state unobservable]}) to explicitly mark the missing phase. This avoids implicit omissions and prevents the model from treating absent mentions as negative evidence. For PFED5+, four task-elicited actions (`Smile', `Frown', `Squeeze eyes', and `Clench teeth') follow the \textit{pre/in/post} template. The `Sit at rest' task does not contain a distinct onset, peak, and offset. We therefore use a single-phase description that corresponds to the \textit{in-action} component, and summarises resting facial state and spontaneous movements over the clip.

\textbf{Clinically Relevant Semantic Content --} The semantic richness of our annotations is derived from fine-grained observations of clinically-relevant facial regions. Under expert guidance, we define which regions should be inspected in each phase.
Annotators then review each clip and record the movement state of each region in each phase. Observations include movement versus no movement, direction of change when visible, and left–right asymmetry when applicable. These structured region-level observations are then assembled into a unified textual description following the standard template.

To maintain clinical faithfulness and reduce hallucination risks during later model training, we adopt a strictly observational annotation policy. Annotators avoid inferential or affective terms and report only what is visually observable. Regions showing no change are explicitly annotated as \textit{no movement} rather than being omitted, which prevents missing mentions from being interpreted as unknown or implicitly active. 
This design ensures that the textual labels function as factual evidence statements rather than severity explanations.

\textbf{Annotation Process and Coverage --}  The textual annotations cover the full PFED5 dataset (2,811 clips). The workflow consists of three stages: (i) Clinicians guide the schema design by defining the regions and observation checklist used for each phase; (ii) Annotators perform manual region-state annotation by recording the motion state of each region per phase for every clip; and (iii) template-based text assembly is performed automatically, where code converts the structured annotations into consistent textual descriptions.

\textbf{Consistency and Quality Assurance --} To improve fluency and reduce redundancy without changing semantics, we apply GPT-4o to post-process the assembled text. The model is provided only with draft textual descriptions. No clinical videos, frames, or identifiable visual data are uploaded for this refinement step and the refinement prompt enforces strict constraints. It forbids adding new facts, forbids inference beyond the provided observations, and requires preserving the meaning of all region-level evidence statements.

All refined descriptions are then fully double-checked and validated by clinical experts to ensure factual correctness and template compliance. 
{Additional details, including the complete template, slot schema, refinement prompt, and annotation statistics, are provided in Supplementary Materials.}


\vspace{-0.8em}
\section{TraMP-LLaMA}
\label{sec6:tramp_llama}

\subsection{Problem Formulation}
This work proposes a joint scoring and reporting solution for video-based FEQA. {Let's consider a} facial video clip $X_{v} \in \mathbb{R}^{T_v \times H \times W \times 3}$ and its corresponding landmark trajectories $X_{m} \in \mathbb{R}^{T_m \times P \times 5}$, where 
{$T_v$ and $T_m$ indicate the frame length and trajectory length, respectively,}
$P=63$ is the number of landmarks and the 5 channels encode $(x,y)$ coordinates {along with the RGB value sampled at each landmark location}. Then,  the model $\mathcal{F}$, where
\begin{equation}
    \hat{s},~ \hat{r} = \mathcal{F}(X_{v}, X_{m}, I)~,
\end{equation}
outputs (i) {a severity score $\hat{s}$ }aligned with clinical ratings (e.g., MDS-UPDRS~\cite{mds-updrs2008}), and (ii) {a template-constrained structured report $\hat{r}$ }
that describes observable facial motor evidence under the template defined in Section~\ref{sec6:text_template}. $I$ denotes the instruction prompt that specifies the reporting format and task context. 
We temporally align the video and trajectory streams such that $T_v$ and $ T_m$ are consistent.
Supervision is provided by {the severity label $s$ and the motion-description label $r$.} The report is intended to externalise evidence that supports the predicted score, rather than to restate the score or provide diagnostic conclusions.

\subsection{Network Architecture}

\begin{figure}[htbp!]
    \centering
    \includegraphics[width=0.95\linewidth]{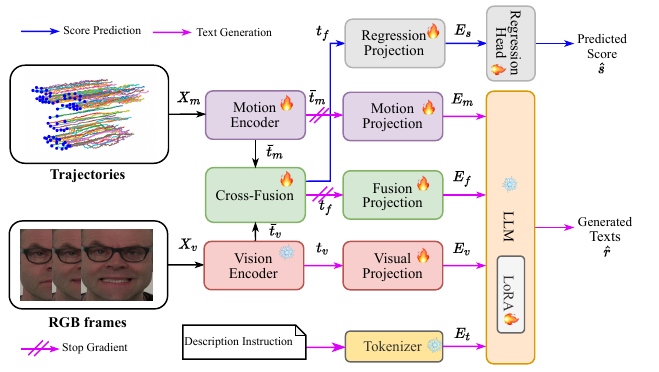}
    \caption{{\bf TraMP-LLaMA for decoupled severity scoring and text generation.} Landmark trajectories are encoded by a motion encoder (SkateFormer~\cite{do2024skateformer}), RGB frames are encoded by a frozen vision encoder in VideoLLaMA3~\cite{zhang2025videollama}, and integrated via cross-fusion. The visual, fused, and motion evidence are projected into the LLM embedding space as $[E_v; E_f; E_m]$. The severity score is predicted by a regression head from $E_s$. For text generation, an instruction prompt tokenised and embedded as $E_t$ conditions a LoRA-adapted LLM together with the evidence embeddings $[E_v; E_f; E_m]$. Gradients from the text-generation loss are blocked from the motion encoder and cross-fusion, which are updated only through the score-prediction objective.}
    \label{fig:tramp_llama}
\end{figure}

Fig.~\ref{fig:tramp_llama} illustrates the proposed TraMP-LLaMA framework. The architecture comprises a trainable trajectory encoder for landmark motion, a frozen vision encoder for RGB frames, a trainable cross-fusion module that integrates motion and appearance into a fused representation, and two decoupled output branches for severity scoring and report generation.

{\bf Visual encoding --} We adopt VideoLLaMA3~\cite{zhang2025videollama}, consisting of SigLIP-NaViT \cite{Zhai2023SigmoidLF} as vision encoder and Qwen-2.5 \cite{Yang2024Qwen25TR} as LLM, as the base VLM due to its strong video understanding capability and its native compatibility with LLM-conditioned token interfaces, which aligns with our design of projecting multimodal evidence into the LLM embedding space. RGB frames $X_v$ are encoded by SigLIP-NaViT, producing a sequence of visual tokens ${t_v}$ that capture appearance and coarse spatio-temporal context. In our implementation, the vision encoder is frozen during training, as indicated in Fig.~\ref{fig:tramp_llama}. The token sequence ${t_v}$ is mapped to the LLM embedding space by the visual projector, yielding the visual token embeddings $E_v \in \mathbb{R}^{T_{v'}\times d}$, where $T_{v'} < T_{v}$ due to token compression used in VideoLLaMA3~\cite{zhang2025videollama} and $d$ denotes the LLM embedding dimension. For cross-fusion, we apply global average pooling over $t_v$ to obtain a clip-level visual embedding $\bar{t}_v$, which serves as the visual input to the cross-fusion module.

{\bf Trajectory encoding --} 
To provide explicit motion evidence, we encode landmark trajectories $X_m$ using a trainable SkateFormer~\cite{do2024skateformer}, following the trajectory representation introduced in \cite{tramp-former}. SkateFormer produces a sequence of motion embeddings which are aggregated by global average pooling to produce a clip-level motion feature $\bar{t}_m$. The motion projector then maps $\bar{t}_m$ to the LLM embedding space to obtain $E_m \in \mathbb{R}^{1 \times d}$. We study the effect of this aggregation choice in Section~\ref{subsec6:ablation}. 

{\bf Cross-fusion representation --} 
The clip-level features $(\bar{t}_m, \bar{t}_v)$ are fed into a trainable cross-fusion module, following \cite{tramp-former}, producing a compact fused representation $t_f$ that aggregates motion-relevant cues. This fused representation is projected to the LLM embedding space to obtain $E_f \in \mathbb{R}^{1 \times d}$ for report generation. In parallel, it is projected to a scoring embedding $E_s \in \mathbb{R}^{1 \times d_r}$ for severity prediction by the scoring branch, where $d_r$ is the input dimension of the regression head.

{\bf Token fusion and LLM conditioning --} 
For report generation, TraMP-LLaMA conditions the LLM on evidence from three sources: the visual token embeddings $E_v$ (RGB appearance context), the trajectory-based motion embedding $E_m$ (explicit motion cues), and the cross-fused embedding $E_f$ (a compact motion-aware summary). We inject $E_m$ and $E_f$ as additional evidence tokens and concatenate them with the visual token sequence $E_v$ to form the multimodal context for the LLM (i.e., $[E_v; E_f; E_m]$). Together with the tokenised instruction prompt $E_t$, the model is instruction-tuned to generate reports. The description instructions for training and inference are detailed in the {Supplementary Materials}.

Compared with conditioning the LLM directly on heterogeneous raw visual and motion token sequences, the cross-fusion pathway distills motion-relevant information into a compact summary under a limited token budget, reducing the burden on the LLM to infer fine-grained motor evidence from long and weakly aligned sequences. In addition, providing $E_m$ alongside $E_f$ helps retain explicit trajectory cues that may be reduced by fusion, enabling more precise evidence statements (e.g., directionality and asymmetry). The contribution of each evidence source ($E_v$, $E_m$, and $E_f$) is ablated in Section~\ref{subsec6:ablation}.

The tokenizer is frozen, and the base LLM is adapted using LoRA \cite{Hu2021LoRALA}. The model generates the report autoregressively conditioned on the instruction and the multi-modal evidence. The visual, motion, and fusion projectors follow the multimodal projector design in VideoLLaMA3~\cite{zhang2025videollama}, implemented as a two-layer MLP with GELU activation. The regression projector is 
a single fully-connected layer.

{\bf Two-head outputs --} TraMP-LLaMA produces two outputs per clip: {a report $\hat{r}$} generated autoregressively by the LLM conditioned on $E_t$ and the evidence embeddings $[E_v; E_f; E_m]$, and {a severity score $\hat{s}$ }
predicted by a regression head from $E_s$. The scoring branch bypasses the LLM.

\subsection{Decoupled Training and Stop-gradient Design}
\label{subsec6:stopgrad}

A core design choice in our approach (as depicted in Fig.~\ref{fig:tramp_llama}) is the decoupling between score learning and language supervision. TraMP-LLaMA is trained with two supervision signals: (i) clinical severity supervision for score prediction ($\mathcal{L}_{score}$), and (ii) text supervision for report generation ($\mathcal{L}_{text}$). Naively propagating both signals through shared spatio-temporal representations can introduce task interference, where $\nabla \mathcal{L}_{text}$ driven by language generation distort the spatio-temporal features required for accurate severity grading. To mitigate this, we adopt a stop-gradient mechanism that decouples the optimisation of the scoring representations from the language-generation objective.

The scoring branch is optimised only by $\mathcal{L}_{score}$, while the report-generation branch is optimised by $\mathcal{L}_{text}$. We  explicitly block $\nabla \mathcal{L}_{text}$ from flowing into the spatio-temporal representations used for scoring, namely the motion encoder output $\bar{t}_m$ and the fused representation $t_f$. That is, we apply stop-gradient at $(\bar{t}_m, t_f)$ with respect to $\mathcal{L}_{text}$, such that the motion encoder and the cross-fusion module are updated only by $\mathcal{L}_{score}$, while detached $\bar{t}_m$ and $t_f$ can still be consumed as evidence by the report-generation branch. 

Meanwhile, severity supervision does not update the LLM parameters: $\nabla \mathcal{L}_{score}$ is not allowed to backpropagate into the LLM. During joint training, $\mathcal{L}_{text}$ updates only the language-side lightweight adaptation components, i.e., LoRA parameters and projection layers (with the vision encoder frozen), whereas $\mathcal{L}_{score}$ updates only the scoring branch, i.e., the motion encoder, cross-fusion module, and regression head. This gradient routing is empirically validated in Table~\ref{tab6:stop_gradient}.

\subsection{Optimisation}
TraMP-LLaMA is trained using a multi-task learning framework with two objectives: clinical severity scoring and structured report generation.
For score prediction, we minimise the MSE loss between the {predicted score $\hat{s}$ and the target clinical rating $s$,} 
\begin{equation}
    \mathcal{L}_{score}=\frac{1}{N}\sum_{i=1}^{N}(\hat{s}_i-s_i)^2~,
\end{equation}
where $N$ denotes the batch size.
For the text generation task, the LLM is fine-tuned via LoRA using the standard autoregressive next-token cross-entropy loss 
\begin{equation}
    \mathcal{L}_{text}=-\frac{1}{N}\sum_{i=1}^{N}\sum_{j=1}^{L_i}\log p_\theta\!\left(t_{i,j}\mid t_{i,<j},E_t,{E}_i\right)~,
\end{equation}
where $t_{i,j}$ denotes the $j^{th}$ ground-truth token of the reference report for sample $i$, $t_{i,<j}$ are the preceding ground-truth tokens, and $L_i$ is the report length. $p_\theta(\cdot)$ is the LLM next-token probability parameterised by $\theta$, conditioned on the instruction prompt embeddings $E_t$ and the multimodal evidence ${E}_i=[E_v;E_f;E_m]$.
During training, we jointly optimise these objectives with the decoupled gradient routing 
as
\begin{equation}
    \mathcal{L}_{total}= \mathcal{L}_{text}+ \lambda \cdot\mathcal{L}_{score}~,
    \label{eq:loss}
\end{equation}
where $\lambda$ balances the two terms. 
We ablate $\lambda$ in Section~\ref{subsec6:ablation}.

\section{Experiments and Results}
\label{sec6:results}

This section evaluates TraMP-LLaMA on {PFED5+} for joint severity scoring and structured report generation. We provide quantitative performance metrics, qualitative analysis of the generated reports, and ablation studies to validate key components such as evidence composition and stop-gradient decoupling.
{We use PFED5+ (as detailed in 
Section~\ref{sec6:text_template}), which includes MDS-UPDRS scores and the structured text annotations.}
Following \cite{duan2023qafe,tramp-former}, 30 subjects are allocated for training and 11 for inference.
Faces are detected and cropped using the SCRFD~\cite{guo2021SCRFD}. 
Video frames are processed at a resolution of $256 \times 256$. Each video clip is annotated with a severity score for quantitative evaluation and a structured report for report-generation evaluation.

{\bf Evaluation Metrics --}
We assess the model using metrics tailored to both severity regression and clinical report generation: for {\it Severity scoring}, 
Spearman’s rank correlation coefficient is reported to measure the monotonic relationship between predicted scores and clinical ratings, following \cite{tramp-former}. For {\it Report generation}, The quality of generated clinical reports is evaluated against expert references using a combination of semantic and n-gram based metrics. For semantic similarity, we report BERTScore (F1)~\cite{Zhang2019BERTScoreET}, which aligns reference and generated tokens via contextual embeddings and cosine similarity. Specifically, given a reference sequence $t=(t_1,\dots,t_m)$ and a generated sequence $\hat{t}=(\hat{t}_1,\dots,\hat{t}_n)$, let $h(\cdot)$ denote the contextual embedding of a token and $c_{ij}=\cos(h(t_i),h(\hat{t}_j))$. BERTScore computes
\begin{equation}
    P_{BERT}=\frac{1}{n}\sum_{j=1}^{n}\max_{i} c_{ij},\quad
R_{BERT}=\frac{1}{m}\sum_{i=1}^{m}\max_{j} c_{ij},\quad
F_1=\frac{2{P_{BERT}}R_{BERT}}{P_{BERT}+R_{BERT}}~.
\end{equation}
We compute BERTScore using DeBERTa-XLarge-MNLI\footnote{https://huggingface.co/microsoft/deberta-xlarge-mnli} as the underlying contextual encoder.
To quantify the overlap of word sequences, we report BLEU-4 \cite{Papineni2002BleuAM} (modified n-gram precision with a brevity penalty) and ROUGE-L \cite{Lin2004ROUGEAP} (longest common subsequence–based overlap). We additionally report CIDEr \cite{Vedantam2014CIDErCI}, which measures term frequency–inverse document frequency weighted n-gram consensus, down-weighting frequent generic phrases and emphasising informative terms. All text metrics are computed on the full generated report under the template, using a single expert reference per clip.

\subsection{Implementation Details}
\label{subsec6:implement}
{\bf Base VLM --}
We build on the VideoLLaMA3-7B Stage-3 fine-tuned checkpoint\footnote{https://huggingface.co/DAMO-NLP-SG/VideoLLaMA3-7B}~\cite{zhang2025videollama}, which uses Qwen2.5-7B~\cite{Yang2024Qwen25TR} as the language backbone with embedding dimension $d=3584$ and a maximum context length of $8192$ tokens. The vision encoder is SigLIP-NaViT~\cite{Zhai2023SigmoidLF} with hidden size $1152$. To reduce the effective visual token length, we apply (i) a video merge size of $2$, which reduces the spatial token grid by a factor of $2$ in each spatial dimension (i.e., $4\times$ fewer tokens per frame) before visual encoding, and (ii) token compression (enabled for both training and inference), which prunes temporally redundant tokens after visual encoding, yielding the processed visual tokens $t_v$  (default threshold $0.1$).

We sample RGB frames at $25$ fps and set the maximum number of frames to $T_v=64$. If a clip contains fewer than $64$ frames, all available frames are used; otherwise, $64$ frames are uniformly sampled. Frames are resized to $224\times224$. To accommodate variable-length clips without temporal padding, SigLIP-NaViT uses NaViT-style~\cite{Dehghani2023PatchNP} token packing with attention masking, packing tokens from multiple clips into a single sequence so each clip is processed at its true length and attention does not cross clip boundaries.

{\bf Motion stream --}
The motion encoder is SkateFormer~\cite{do2024skateformer} pretrained on landmark trajectories from DFEW~\cite{DFEW}. Following \cite{tramp-former}, the SkateFormer encoder consists of four blocks with dimensions $(128, 256, 256, 256)$, respectively.  We set the temporal length to $T_m=T_v=64$ and apply zero padding when a clip contains fewer than 64 frames, as SkateFormer requires fixed-length trajectory inputs. We apply random clip reversal as motion augmentation during training.

{\bf Fusion and projection layers --} We follow \cite{tramp-former} to use three cross-fusion blocks, each producing the fused features with the dimension of 256. The visual projector follows the VideoLLaMA3 design, implemented as a two-layer MLP with a GELU non-linearity in between, mapping visual token features to the LLM embedding space (i.e., $1152 \rightarrow 3584 \rightarrow 3584$). The motion and fusion projectors use the same MLP design
to map the features with the dimension of 256 to the LLM space (i.e., $256 \rightarrow 3584 \rightarrow 3584$).
To interface the visual stream with the fusion module, we apply a linear video-to-fusion projection ($1152\rightarrow256$).
For severity scoring, the regression branch consists of a linear layer ($256\rightarrow256$) followed by the final linear head ($256\rightarrow1$).

{\bf Training regime --}
LoRA~\cite{Hu2021LoRALA} is used to adapt the LLM with rank $\mathrm{lora}_r=64$, $\mathrm{lora}_\alpha=128$, and dropout $0.05$. LoRA is applied to attention projections $(q, k, v, o)$ and MLP projections $(up, down)$ in each Transformer layer. Module-wise learning rates are set to $5\times10^{-6}$ for LoRA parameters and the visual projector, and $5\times10^{-5}$ for the remaining trainable components, including the video-to-fusion projector, motion projector, fusion projector, fusion modules, and the regression projector/head.

All trainable parameters are optimised with AdamW (weight decay $0.01$) using a cosine learning-rate schedule with a warmup ratio of $0.03$. Training runs for 6 epochs with a per-GPU batch size of 1 on 4 NVIDIA GH200 GPUs. We do not use gradient accumulation (accumulation steps $= 1$), resulting in an effective batch size of 4.
Mixed-precision training uses BF16, and gradient checkpointing is enabled for memory efficiency. DeepSpeed ZeRO-2~\cite{Rajbhandari2019ZeROMO} is used to partition optimiser states and gradients across data-parallel workers.

For scoring, clinical severity labels are normalised from $[0,4]$ to $[0,1]$ during training (i.e., $y \leftarrow y/4$) and rescaled back at inference. The overall objective weights the regression term (i.e., MSE loss) by $\lambda=5$.  We use SDPA attention for compatibility on the aarch64-based GH200 environment.

\subsection{Comparative Results}
\label{subsec6:compare}

We evaluate TraMP-LLaMA on {PFED5+} for both severity scoring and report generation. Since TraMP-LLaMA is built upon VideoLLaMA3 \cite{zhang2025videollama}, the most direct comparison is with this architecturally matched baseline.  For severity scoring, we further compare against representative DFER baselines (Former-DFER~\cite{formerDFER} and S2D~\cite{chen2024static}) and AQA baselines (USDL~\cite{2020USDL}, CoFInAl~\cite{zhou2024cofinal}, QAFE-Net~\cite{duan2023qafe}, and TraMP-Former~\cite{tramp-former}) as broader reference methods. For report generation, we additionally include recent general-purpose VLMs, including Chat-UniVi~\cite{jin2024chat}, LLaVA-Video~\cite{zhang2025llavavideo}, and LongVU~\cite{Shen2024LongVUSA}. These models are used only for report-generation comparison, as they are not designed for score regression.

Unlike previous action quality assessment works, where models are trained and evaluated per expression, we train a single model jointly across all {PFED5+} actions to better reflect the target clinical setting. 
DFER baselines are converted to regressors by replacing the classification head with a three-layer MLP regression head and initialised from their DFEW fine-tuned checkpoints. AQA baselines are initialised from their publicly released pretrained weights. All scoring baselines are then fine-tuned on {PFED5+} under the same training protocol. To align with the joint-action training setting, we add an auxiliary cross-entropy objective for predicting the {PFED5+} action identity alongside the primary regression loss, weighted equally.

For report-generation baselines, each VLM is initialised from its publicly released fine-tuned checkpoint. 
For fair comparison, the vision encoder is kept frozen, while the multimodal projector and the LoRA parameters of the LLM remain trainable. All report-generation baselines are fine-tuned using the same instruction format, frame sampling strategy, and decoding settings unless otherwise specified. In all result tables, the best result is highlighted in \textbf{bold} and the second-best is \underline{underlined}. All metrics are reported in ($\%$) except CIDEr~\cite{Vedantam2014CIDErCI}, which is reported on its original scale because it is non-negative and not upper-bounded, making percentage-style scaling less interpretable.

{\bf Score Prediction --} Table~\ref{tab6:pfed5_score} reports the comparative results of severity scoring on {PFED5+} across five clinical actions. TraMP-LLaMA achieves the best performance, obtaining $57.35\%$ in average correlation across actions and $56.63\%$ in all-action correlation, outperforming all adapted DFER and AQA baselines, as well as the matched VLM baseline VideoLLaMA3.

Compared with QAFE-Net (Avg. $53.11\%$) and TraMP-Former (Avg. $45.04\%$), TraMP-LLaMA achieves higher average correlation, with gains of $\uparrow4.24\%$ and $\uparrow12.31\%$, respectively. These comparisons indicate that the proposed framework remains competitive against representative quality-assessment baselines even under the more challenging joint-action setting.
Importantly, compared with the same base VLM (VideoLLaMA3), adding trajectory cues with cross-fusion improves the average correlation by $\uparrow2.29\%$ and the overall correlation by $\uparrow5.14\%$. The gains are most evident on `Sit at rest' ($\uparrow10.40\%$), `Frown' ($\uparrow6.83\%$), and `Squeeze eyes' ($\uparrow9.76\%$), indicating that trajectory cues are especially helpful for these actions in the joint-action setting.

For `Smile' and `Clench teeth', TraMP-LLaMA underperforms VideoLLaMA3. This may reflect ambiguity in the mouth region, where subtle smiles and mild clenching can appear visually similar under limited teeth visibility or partial occlusion. In such cases, trajectory cues may be less reliable or less discriminative, especially under joint-action training.

\begin{table}[t]
\begin{center}
\resizebox{\columnwidth}{!}{
\begin{tabular}{lll|ccccccc} 
\toprule
\textbf{Methods} & \textbf{Publication} & \begin{tabular}[c]{@{}l@{}}\textbf{Vision}\\\textbf{Backbone}\end{tabular} & \begin{tabular}[c]{@{}c@{}}\textbf{Sit }\\\textbf{at rest}\end{tabular} & \textbf{Smile} & \textbf{Frown} & \begin{tabular}[c]{@{}c@{}}\textbf{Squeeze }\\\textbf{eyes}\end{tabular} & \begin{tabular}[c]{@{}c@{}}\textbf{Clench }\\\textbf{teeth}\end{tabular} & \begin{tabular}[c]{@{}c@{}}\textbf{Avg.}\\\textbf{Corr.}\end{tabular} & \begin{tabular}[c]{@{}c@{}}\textbf{All}\\\textbf{Corr.}\end{tabular} \\ 
\hline
Former-DFER \cite{formerDFER} & MM'21 & CNN-ViT & 46.86 & 44.27 & 42.42 & 51.94 & 43.31 & 45.76 & 46.15 \\
S2D \cite{chen2024static} & TAFFC'24 & \begin{tabular}[c]{@{}l@{}}ViT\\+MFN\end{tabular} & 35.57 &48.18 &52.21&38.94&38.04&42.59 &42.36\\ 
\hline
USDL \cite{2020USDL} & CVPR'20 & I3D & 28.06 & 51.85 &38.84  &44.91  &\underline{49.67}  & 42.67 &40.25  \\
CoFInAl \cite{zhou2024cofinal} & IJCAI'24 & I3D & 36.05 &25.92&12.62&12.00&12.06&19.73& 20.32 \\
CoFInAl \cite{zhou2024cofinal} & IJCAI'24 & VST & 26.64&28.88&25.91&16.65&24.25&24.66& 24.70 \\
QAFE-Net \cite{duan2023qafe} & WACVW'24 & \begin{tabular}[c]{@{}l@{}}CNN-ViT\\+SlowOnly\end{tabular} & \underline{50.96} & 48.01 & 53.70 & \underline{60.69} & \textbf{52.20} & 53.11 & \underline{52.24} \\
TraMP-Former \cite{tramp-former} & FG'25 & \begin{tabular}[c]{@{}l@{}}CNN-ViT\\+SkateFormer\end{tabular} & 34.02 & 38.89 & 54.36 & 59.64 & 38.28 & 45.04 & 44.30 \\ 
\hline
VideoLLaMA3 \cite{zhang2025videollama} & ArXiv'25 & SigLIP-NaViT & {50.89} & \textbf{62.45} & \underline{61.85} & 53.67 & {46.43} & \underline{55.06} & 51.49 \\
\begin{tabular}[c]{@{}l@{}}TraMP-LLaMA\\(Ours)\end{tabular} & - & \begin{tabular}[c]{@{}l@{}}SigLIP-NaViT\\+SkateFormer\end{tabular} & \textbf{61.29} & 55.36 & \textbf{68.68} & \textbf{63.43} & 38.00 & \textbf{57.35} & \textbf{56.63} \\
\bottomrule
\end{tabular}}
\end{center}
\caption{{\bf Severity scoring results on PFED5+ across five clinical actions.} MFN: MobileFaceNet.}
\label{tab6:pfed5_score}
\end{table}

{\bf Report Generation --} 
Table~\ref{tab6:pfed5_report} compares structured report generation performance on {PFED5+} against VLM baselines. The most direct comparison is with VideoLLaMA3, since TraMP-LLaMA is built on the same backbone and uses the same LLM (i.e., Qwen2.5-7B).
Under this matched setting, TraMP-LLaMA achieves better results across all metrics, including 67.39\% on BERTScore, improving over VideoLLaMA3. 
Although the absolute differences are modest, the consistent gains across all metrics suggest that the proposed motion-aware design helps improve report generation quality, yielding slightly better alignment with expert references across both semantic-similarity and n-gram–based metrics.

Compared with other generic VLMs, both TraMP-LLaMA and VideoLLaMA3 perform substantially better. This gap may reflect both task mismatch and differences in backbone capability. PFED5+ requires the model to capture subtle, region-level facial motion and express it in a structured report format, which may be less well aligned with VLMs trained mainly for broad video understanding. VideoLLaMA3 provides a stronger starting point in this setting, and TraMP-LLaMA further improves on it by introducing explicit trajectory-based motion cues.
Nevertheless, the improvements are small, indicating substantial room to strengthen report reliability. In particular, format deviations (e.g., missing required stages), empty outputs, and unsupported or incorrect region statements may still be observed in a small number of cases, which can limit clinical usability even when overlap-based metrics improve.


\begin{table}[htbp!]
\centering
\resizebox{\columnwidth}{!}{
\begin{tabular}{llllcccc} 
\toprule
\textbf{Methods} & \textbf{Publication} & \begin{tabular}[c]{@{}l@{}}\textbf{Vision}\\\textbf{Encoder}\end{tabular} & \textbf{LLM} & \begin{tabular}[c]{@{}c@{}}\textbf{BERT}\\\textbf{Score}\end{tabular} & \textbf{ROUGE-L} & \textbf{BLEU-4} & \textbf{CIDEr} \\ 
\hline
Chat-UniVi \cite{jin2024chat} & CVPR'24 & CLIP ViT-L/14~ & Vicuna-7B & 49.13 & 21.33 & 3.26 & 0.026 \\
LLaVA-Video \cite{zhang2025llavavideo} & TMLR'25 & SigLIP-so400m~ & Qwen2-7B & 35.16 & 18.85 & 1.98 & 0.016 \\
LongVU~\cite{Shen2024LongVUSA} & ICML'25 & SigLIP-so400m & Qwen2-7B & 47.65 & 20.64 & 2.26 & 0.022 \\ 
\hline
VideoLLaMA3 \cite{zhang2025videollama} & ArXiv'25 & SigLIP-NaViT & Qwen2.5-7B & \underline{66.89} & \underline{40.93} & \underline{27.25} & \underline{0.2632} \\
TraMP-LLaMA (Ours) & - & \begin{tabular}[c]{@{}l@{}}SigLIP-NaViT\\+SkateFormer\end{tabular} & Qwen2.5-7B & \textbf{67.39} & \textbf{41.88} & \textbf{28.00} & \textbf{0.2847} \\
\bottomrule
\end{tabular}}
\caption{{\bf Comparison of structured report generation performance on PFED5+ against VLM baselines.}}
\label{tab6:pfed5_report}
\end{table}

\subsection{Ablations}
\label{subsec6:ablation}

We ablate key design choices in TraMP-LLaMA, including loss balancing, evidence composition for LLM conditioning, temporal fusion design, and the stop-gradient decoupled training strategy. Unless stated otherwise, all ablations follow the same training and evaluation protocol as in Section~\ref{subsec6:implement}.

\textbf{Impact of Loss Balancing --} We vary $\lambda\in\{1,3,5,7,10\}$ in Eq.~\ref{eq:loss} (with the stop-gradient routing fixed as described in Section~\ref{subsec6:stopgrad}) to balance the regression and generation objectives. Here, $\mathcal{L}_{score}$ is computed on normalised labels in $[0,1]$. As shown in Table~\ref{tab6:lambda}, $\lambda=5$ provides the best overall trade-off, achieving the highest overall correlation ($56.63\%$), BERTScore ($67.39\%$), and CIDEr ($0.2847$), while maintaining competitive performance on the remaining metrics.
We therefore use $\lambda=5$ as the default setting in all subsequent experiments.

\begin{table}[htbp!]
\begin{center}
\resizebox{\columnwidth}{!}{
\begin{tabular}{c|ccccccc|cccc} 
\toprule
\multirow{3}{*}{\textbf{$\lambda$}} & \multicolumn{7}{c|}{\textbf{\textit{Score Prediction}}} & \multicolumn{4}{c}{\textbf{\textit{Report Generation}}} \\
 & \begin{tabular}[c]{@{}c@{}}\textbf{Sit }\\\textbf{at rest}\end{tabular} & \textbf{Smile} & \textbf{Frown} & \begin{tabular}[c]{@{}c@{}}\textbf{Squeeze }\\\textbf{eyes}\end{tabular} & \begin{tabular}[c]{@{}c@{}}\textbf{Clench }\\\textbf{teeth}\end{tabular} & \begin{tabular}[c]{@{}c@{}}\textbf{Avg.}\\\textbf{Corr.}\end{tabular} & \begin{tabular}[c]{@{}c@{}}\textbf{All}\\\textbf{Corr.}\end{tabular} & \begin{tabular}[c]{@{}c@{}}\textbf{BERT}\\\textbf{Score}\end{tabular} & \textbf{ROUGE-L} & \textbf{BLEU-4} & \textbf{CIDEr} \\ 
\hline
1 & \textbf{62.05} & \textbf{62.44} & 63.33 & 54.90 & \textbf{ 44.22 } & \textbf{57.39} & \underline{55.80} & 66.62 & \textbf{43.49} & \textbf{28.35} & 0.2549 \\
3 & 61.12 & 54.07 & 66.28 & 60.47 & 36.10 & 55.61 & 54.45 & 67.07 & 41.30 & 27.70 & 0.2778 \\
5 & \underline{61.29} & \underline{ 55.36 } & \textbf{68.68} & \textbf{63.43} & \underline{ 38.00 } & \underline{57.35} & \textbf{56.63} & \textbf{67.39} & \underline{41.88} & \underline{28.00} & \textbf{0.2847} \\
7 & 61.09 & 51.61 & 65.70 & 60.44 & 32.18 & 54.20 & 53.41 & \underline{67.24} & 41.46 & 27.76 & \underline{0.2810} \\
10 & 55.90 & 54.23 & \underline{67.86} & \underline{61.94} & 33.32 & 54.65 & 54.44 & 66.34 & 39.90 & 26.13 & 0.2612 \\
\bottomrule
\end{tabular}}
\end{center}
\caption{{\bf Comparison of different loss weights $\lambda$ for balancing score prediction and report generation on {PFED5+}.}}
\label{tab6:lambda}
\end{table}

\textbf{Gradient Flow Analysis --}
Table~\ref{tab6:stop_gradient} shows the ablation of the gradient-routing choice within the decoupled training scheme (as detailed in Section~\ref{subsec6:stopgrad}). Specifically, we test whether the text-generation objective $\mathcal{L}_{text}$ should be allowed to backpropagate into the motion-aware representations used for scoring, namely the motion encoder output $\bar{t}_m$ and the fused representation $t_f$ in Fig.~\ref{fig:tramp_llama}, thereby updating the motion encoder and cross-fusion module, or whether these gradients should be blocked as in our stop-gradient design.
We compare two settings while keeping all other configurations identical: (i) enabling $\nabla \mathcal{L}_{text}$ to update the motion encoder and cross-fusion module, and (ii) blocking $\nabla \mathcal{L}_{text}$ at $(\bar{t}_m, t_f)$ (ours). Table~\ref{tab6:stop_gradient} shows that blocking text gradients yields substantially higher scoring correlation and also improves report quality, confirming the benefit of the proposed stop-gradient routing for stable severity grading. We additionally attempted naive joint optimisation without stop-gradient routing; under the same training setup, optimisation was highly unstable and did not yield reproducible convergent runs.

\begin{table}[htbp!]
\begin{center}
\resizebox{\columnwidth}{!}{
\begin{tabular}{c|ccccccc|cccc} 
\toprule
\multirow{3}{*}{\begin{tabular}[c]{@{}c@{}}\textbf{$\nabla \mathcal{L}_{text}$}\\\textbf{$\rightarrow(\bar{t}_m,t_f)$}\end{tabular} } & \multicolumn{7}{c|}{\textbf{\textit{Score Prediction}}} & \multicolumn{4}{c}{\textbf{\textit{Report Generation}}} \\
 & \begin{tabular}[c]{@{}c@{}}\textbf{Sit }\\\textbf{at rest}\end{tabular} & \textbf{Smile} & \textbf{Frown} & \begin{tabular}[c]{@{}c@{}}\textbf{Squeeze }\\\textbf{eyes}\end{tabular} & \begin{tabular}[c]{@{}c@{}}\textbf{Clench }\\\textbf{teeth}\end{tabular} & \begin{tabular}[c]{@{}c@{}}\textbf{Avg.}\\\textbf{Corr.}\end{tabular} & \begin{tabular}[c]{@{}c@{}}\textbf{All}\\\textbf{Corr.}\end{tabular} & \begin{tabular}[c]{@{}c@{}}\textbf{BERT}\\\textbf{Score}\end{tabular} & \textbf{ROUGE-L} & \textbf{BLEU-4} & \textbf{CIDEr} \\ 
\hline
Enabled& 39.91&50.56&56.12&36.11&34.96&43.53&44.05&55.73&28.33&13.22&0.0139
\\
Blocked & \textbf{61.29} & \textbf{ 55.36 } & \textbf{68.68} & \textbf{63.43} & \textbf{ 38.00 } & \textbf{57.35} & \textbf{56.63} & \textbf{67.39} & \textbf{41.88} & \textbf{28.00} & \textbf{0.2847} \\
\bottomrule
\end{tabular}}
\end{center}
\caption{{\bf Ablation on stop-gradient routing for motion and fusion representations on {PFED5+}.} We compare enabling vs. blocking gradients from the text-generation loss $\mathcal{L}_{text}$ at $(\bar{t}_m, t_f)$, which respectively allows or prevents $\mathcal{L}_{text}$ from updating the motion encoder and cross-fusion module (Fig.~\ref{fig:tramp_llama}).}
\label{tab6:stop_gradient}
\end{table}

\textbf{Effectiveness of Conditioning Tokens --}
We study whether injecting the trajectory conditioning token $E_m$ and the fused motion summary token $E_f$ into the LLM input is redundant or beneficial  beyond visual tokens $E_v$. As can be seen in Table~\ref{tab6:evidence_features}, using all conditioning tokens ($E_v+E_m+E_f$) achieves the best BERTScore and CIDEr, indicating improved semantic alignment and informative phrase matching. Interestingly, excluding $E_f$ yields higher ROUGE-L and BLEU-4, but lower BERTScore and CIDEr, suggesting a trade-off between surface-form overlap and semantic matching. Since $E_m$ and $E_f$ are each injected as a single token, their overhead is negligible relative to the visual token sequence. Moreover, due to stop-gradient routing, adding these tokens does not affect the scoring pathway (Table~\ref{tab6:stop_gradient}). We therefore retain both $E_m$ and $E_f$ in the default setup.

\begin{table}[htbp!]
\begin{center}
\begin{tabular}{l|cccc} 
\toprule
\begin{tabular}[c]{@{}c@{}}\textbf{LLM}\\\textbf{Conditioning Tokens}\end{tabular} & \begin{tabular}[c]{@{}c@{}}\textbf{BERT}\\\textbf{Score}\end{tabular} & \textbf{ROUGE-L} & \textbf{BLEU-4} & \textbf{CIDEr} \\ 
\hline
$E_v$ only (VideoLLaMA3) & 66.89 & 40.93 & 27.25 & 0.2632 \\
$E_v+E_m$ (w/o $E_f$) & 66.51 & \textbf{43.42} & \textbf{28.49} & 0.2637 \\
$E_v+E_f$ (w/o $E_m$)& \underline{67.05} & 41.31 & 27.57 & \underline{0.2752} \\
$E_v+E_m+E_f$  & \textbf{67.39} & \underline{41.88} & \underline{28.00} & \textbf{0.2847} \\
\bottomrule
\end{tabular}
\end{center}
\caption{{\bf Ablation on conditioning tokens provided to the LLM for report generation on {PFED5+}.} We vary the inclusion of the visual token sequence $E_v$, trajectory conditioning token $E_m$, and fused conditioning token $E_f$.}
\label{tab6:evidence_features}
\end{table}

\textbf{ Temporal Information Aggregation --}
We ablate the temporal form of the fused representation by comparing a frame-wise fusion-feature sequence against global average pooling (GAP). To isolate the effect of temporal aggregation in the fusion pathway, we fix the LLM evidence to $E_v+E_f$ and exclude the landmark token $E_m$. As shown in Table~\ref{tab6:pooling}, GAP yields higher overall scoring correlation and slightly better text metrics. This suggests that pooling can suppress frame-level jitter and provide a more compact conditioning signal under a limited context budget. Notably, retaining the full sequence benefits `Smile' and `Clench teeth', which may indicate that fine-grained temporal dynamics help disambiguate visually similar lower-face actions. Again, since the stop-gradient routing, the score-prediction results under the $E_v+E_f$ setting are consistent with those reported in Table~\ref{tab6:stop_gradient}.

\begin{table}[htbp!]
\begin{center}
\resizebox{\columnwidth}{!}{
\begin{tabular}{c|ccccccc|cccc} 
\toprule
\multirow{3}{*}{\textbf{Fusion $t_f$}} & \multicolumn{7}{c|}{\textbf{\textit{Score Prediction}}} & \multicolumn{4}{c}{\textbf{\textit{Text Generation}}} \\
 & \begin{tabular}[c]{@{}c@{}}\textbf{Sit }\\\textbf{at rest}\end{tabular} & \textbf{Smile} & \textbf{Frown} & \begin{tabular}[c]{@{}c@{}}\textbf{Squeeze }\\\textbf{eyes}\end{tabular} & \begin{tabular}[c]{@{}c@{}}\textbf{Clench }\\\textbf{teeth}\end{tabular} & \begin{tabular}[c]{@{}c@{}}\textbf{Avg.}\\\textbf{Corr.}\end{tabular} & \begin{tabular}[c]{@{}c@{}}\textbf{All}\\\textbf{Corr.}\end{tabular} & \begin{tabular}[c]{@{}c@{}}\textbf{BERT}\\\textbf{Score}\end{tabular} & \textbf{ROUGE-L} & \textbf{BLEU-4} & \textbf{CIDEr} \\ 
\hline
sequence & 49.28 & \textbf{68.06} & 51.23 & 45.92 & \textbf{48.77} & 52.65 & 51.56 & 66.95 & 40.93 & 27.31 & 0.2583 \\
GAP & \textbf{61.29} & 55.36 & \textbf{68.68} & \textbf{63.43} & 38.00 & \textbf{57.35} & \textbf{56.63} & \textbf{67.05} & \textbf{41.31} & \textbf{27.57} & \textbf{0.2752} \\

\bottomrule
\end{tabular}}
\end{center}
\caption{{\bf Ablation on temporal aggregation of fusion features.} We compare retaining the frame-wise fused sequence versus global average pooling (without landmark evidence, i.e., $E_v+E_f$).}
\label{tab6:pooling}
\end{table}

\section{Conclusions}
\label{sec6:conclusion}

This work studied a joint scoring and reporting method for video-based, fine-grained FEQA. The proposed TraMP-LLaMA model is thus a unified framework that predicts clinically aligned severity scores while generating structured, evidence-based motion reports. {To support this evidence-based report generation, we extended PFED5 with expert-guided structured motion descriptions, creating PFED5+.} The model leverages a pretrained VLM backbone with instruction tuning for report generation and incorporates trajectory-based motion cues together with a cross-fusion pathway to support fine-grained assessment under clinically realistic joint-action training.
Experiments on {PFED5+} demonstrated that TraMP-LLaMA achieves SOTA performance for severity scoring and improves report generation quality over strong VLM baselines. 
Ablation studies further support the effectiveness of the proposed design choices, particularly the evidence-token formulation and the decoupled stop-gradient training scheme for stable joint optimisation.

Although this framework can provide interpretability for FEQA, the joint scoring and reporting approach  still has two practical limitations. First, as a generative VLM, TraMP-LLaMA may occasionally hallucinate region-specific details, especially when motion cues are subtle or partially occluded. Second, {PFED5+} remains relatively small and imbalanced across severity levels, and report supervision is constrained by single-reference annotations and inter-rater variability. Broader data collection and stronger supervision, for example multi-reference reports or temporally localised annotation, could further improve both measuring reliability and interpretability.


\bibliography{ref}


\appendix

\section*{Supplementary Materials}

\addcontentsline{toc}{section}{Supplementary Materials}

\setcounter{figure}{0}

\renewcommand{\thefigure}{S\arabic{figure}}

\setcounter{table}{0}

\renewcommand{\thetable}{S\arabic{table}}

\setcounter{equation}{0}

\renewcommand{\theequation}{S\arabic{equation}}

This supplementary materials provides additional details on the text annotations in the PFED5+ dataset, including the exact templates, slot schema, and refinement prompts used to construct the motion-description labels, as well as the description instructions used for training and inference.

\section{Qualitative Comparison of Generated Reports}

\begin{figure}[htbp!]
    \centering
    \includegraphics[width=0.90\linewidth]{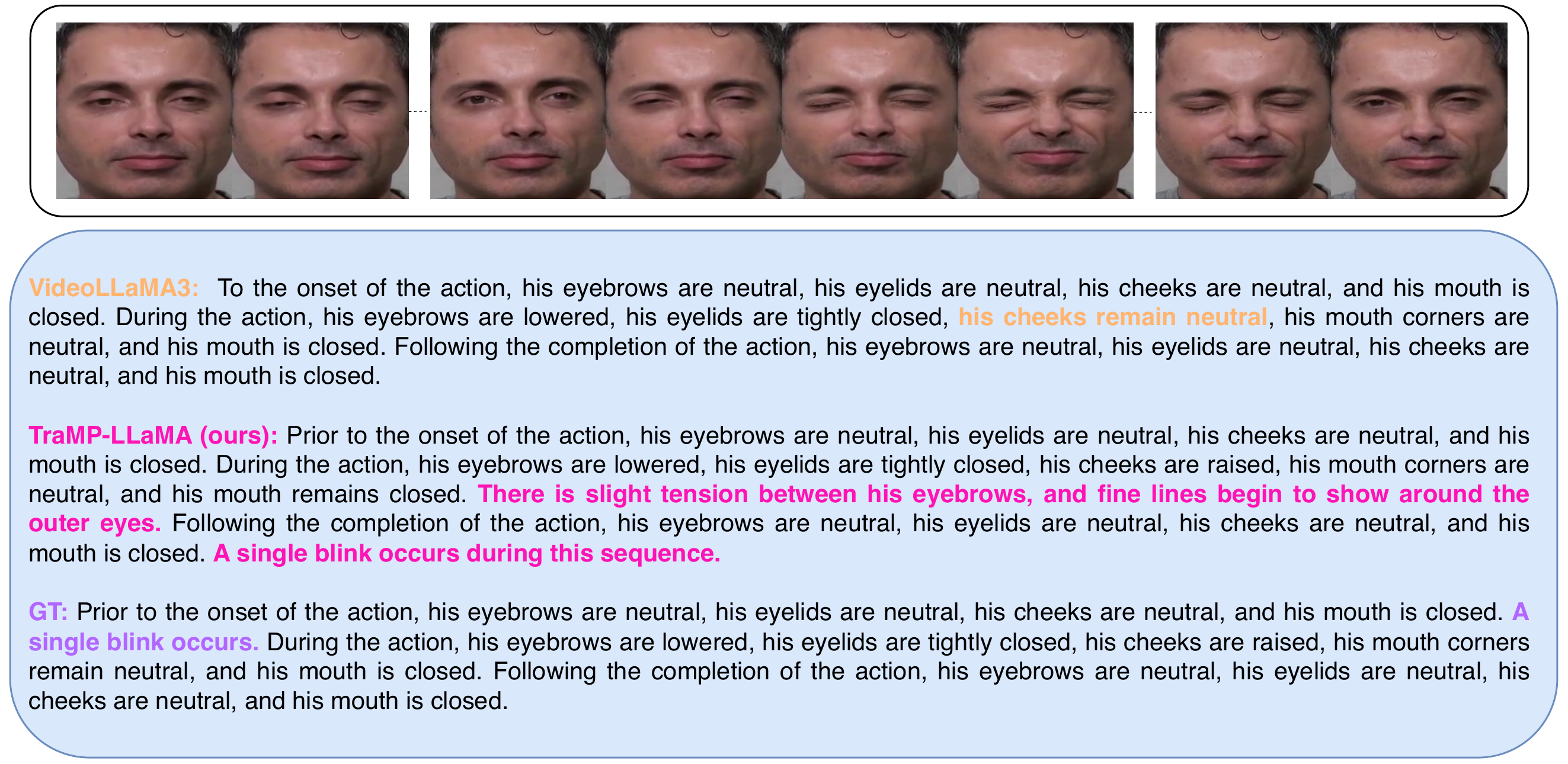}
    \caption{{\bf Qualitative comparison of generated reports for an example of `Squeeze eyes' {in PFED5+}. } VideoLLaMA3 produces an incorrect cheek-state description, whereas TraMP-LLaMA includes additional fine-grained motion cues beyond those explicitly mentioned in the ground-truth report.}
    \label{fig6:case_study_003}
\end{figure}

\begin{figure}[htbp!]
    \centering
    \includegraphics[width=0.90\linewidth]{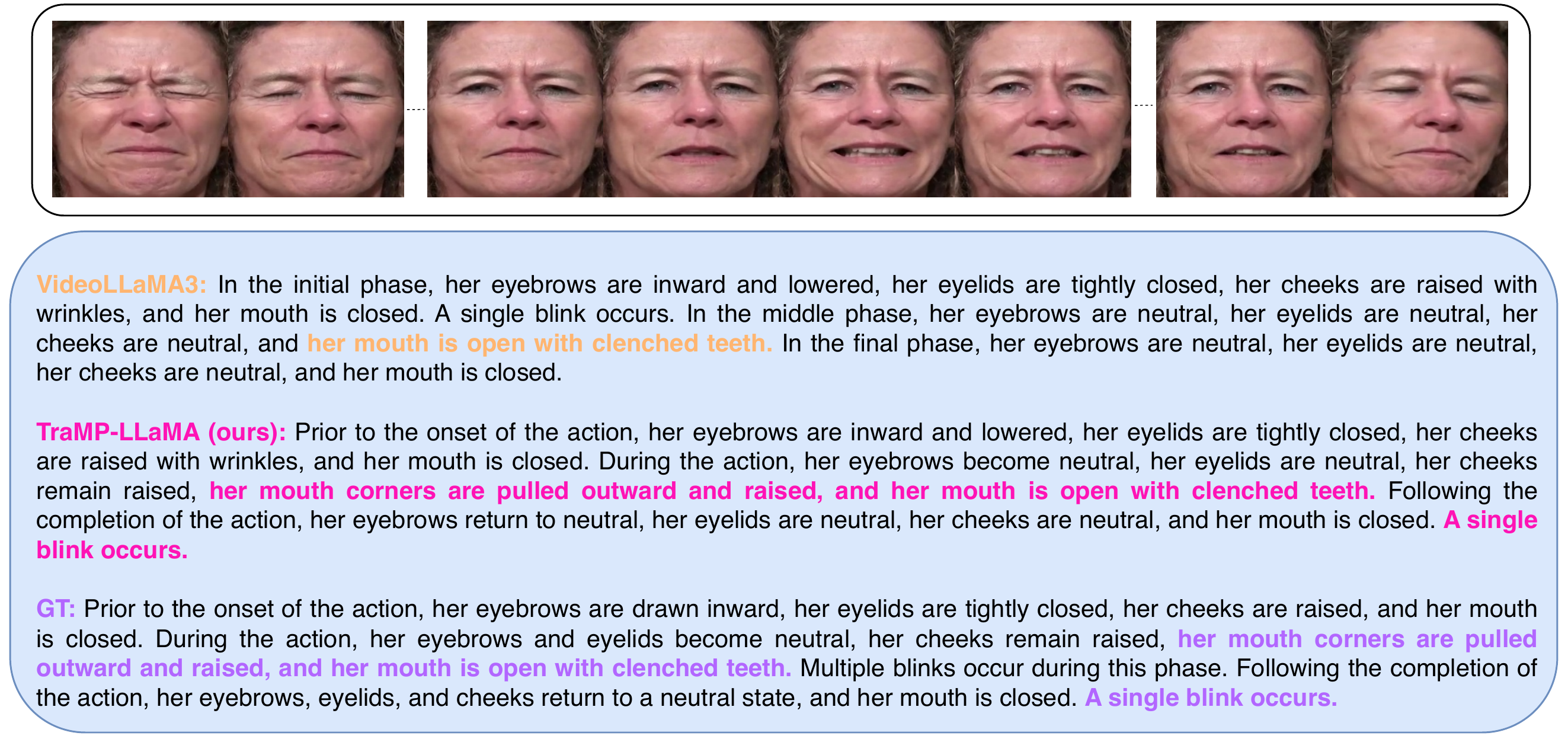}
    \caption{{\bf Qualitative comparison of generated reports for an example of `Clench teeth' {in PFED5+}.}  TraMP-LLaMA’s description is closer to the ground-truth report, particularly for the mouth configuration during the action stage.}
    \label{fig6:case_study_005}
\end{figure}

Here, we present qualitative visualisations of generated reports of open-source baseline and our proposed method in Figs. \ref{fig6:case_study_003} and \ref{fig6:case_study_005}. For each example, we show the Ground Truth report alongside outputs from VideoLLaMA3 and TraMP-LLaMA (ours). To keep the comparison concise, we only highlight the differing or most informative parts. Colours identify different methods, and the same colours are used to mark the corresponding key differences.

Despite generally producing well-structured reports, a small number of outputs exhibit formatting and content issues. In particular, the model may omit descriptions for certain facial regions (e.g., missing eyebrows or eyelids while describing only the mouth or cheeks) and occasionally generate incorrect region-specific details that are not supported by the visual evidence.

\section{Text Annotation Details for PFED5+}


\subsection{Full Templates}
\label{app:full_template}

For task-elicited action clips (`smile', `frown', `squeeze eyes', `clench teeth'), each description follows the same three-phase structure:

\begin{quote}
\textbf{Action template (three-phase).} \\
\textit{Prior to the onset of the action, [pre-action observations]. During the action, [in-action observations]. Following the completion of the action, [post-action observations].}
\end{quote}

The `sit at rest' task does not contain a distinct onset, peak, and offset. It is therefore described using a single-phase template that corresponds to the \texttt{in-action} component:

\begin{quote}
\textbf{Sit-at-rest template (single-phase).} \\
\textit{During this time, [resting-state observations].}
\end{quote}

\subsection{Local-region Structured Annotation Schema}
\label{app:local_region}

Before generating the final paragraph-level text, each clip is first represented as a structured record containing (i) phase keys and (ii) region-state pairs. 

\subsubsection{Region slots}

The structured record uses region-specific slots that reflect clinically relevant facial areas. In PFED5+, the following slots are used:

\begin{itemize}
    \item \textbf{Pre-action/Post-action}: \texttt{Eyebrows}, \texttt{Cheeks}, \texttt{Mouth}, and optional global indicators such as \texttt{Blink} when visible
    \item \textbf{In-action}: \texttt{Eyebrows}, \texttt{Eyelids}, \texttt{Cheeks}, \texttt{Mouth Corners}, \texttt{Mouth}
\end{itemize}

\subsubsection{Example for `Smile'}

\begin{quote}
\textbf{Pre-action:} \{Eyebrows: Neutral; Cheeks: Neutral; Mouth: Closed\} \\
\textbf{In-action:} \{Eyebrows: Neutral; Eyelids: Neutral; Cheeks: Raised with wrinkles; Mouth Corners: Pulled outward and raised; Mouth: Open showing teeth\} \\
\textbf{Post-action:} \{Blink: Once; Eyebrows: Neutral; Cheeks: Neutral; Mouth: Closed\}
\end{quote}

\subsection{Annotation Statistics} 
We summarise the textual labels to characterise annotation density and consistency, including average description length in words, phase-level description length for \texttt{pre/in/post} components, and the prevalence of explicit \textit{no movement} statements and \textit{[state unobservable]} placeholders. Results are reported in Table~\ref{tab:annotation_statistic}.

\begin{table}[htbp!]
\begin{center}
\resizebox{\linewidth}{!}{
\begin{tabular}{lcccccc} 
\toprule
\textbf{Split} & \textbf{\#Clips} & \begin{tabular}[c]{@{}c@{}}\textbf{Avg.}\\\textbf{Words}\end{tabular} & \textbf{\# 3-phase clips} & \begin{tabular}[c]{@{}c@{}}\textbf{Avg. Words }\\\textbf{per Phase}\\\textbf{(\texttt{Pre/In/Post})}\end{tabular} & \begin{tabular}[c]{@{}c@{}}\textbf{\% clips with}\\\textbf{\textit{no movement}}\end{tabular} & \begin{tabular}[c]{@{}c@{}}\textbf{\% clips with }\\\textbf{\textit{[state unobservable]}}\end{tabular} \\ 
\midrule
All & 2811 & 67.14 & 2247 (79.9\%) & 17.07/28.97/16.93 & 5.19\% & 11.03\% \\
Action & 2247 & 78.98 & 2247 (100\%) & 17.07/28.97/16.93 & 6.50\% & 13.80\% \\
Sit at Rest & 564 & 19.99 & 0 & \textemdash & 0\% & 0\% \\
\bottomrule
\end{tabular}}
\end{center}
\caption{{\bf Text annotation statistics of {PFED5+}.}  `3-phase clips' correspond to action tasks using the \texttt{pre/in/post} template; `Sit at rest' uses a single-phase template.}
\label{tab:annotation_statistic}
\end{table}

\subsection{GPT-4o Refinement Prompt}
\label{app:gpt_refine}

\begin{quote}
You are a language expert refining structured facial expression descriptions for a dataset.

Your task is to improve grammatical fluency and remove redundancy while strictly preserving all factual details and the original order of observations.

Constraints:

1) Do not add any new observations or infer any hidden states (e.g., intent, affect, diagnosis).

2) Do not remove any information about facial details (eyebrows, eyelids, cheeks, mouth corners, mouth).

3) Keep the same template structure.
   - For `sit at rest', keep the format:
     ``During this time, ...''
   - For action clips (`smile', `frown', `squeeze eyes', `clench teeth'), keep the format:
     ``Prior to the onset of the action, ... During the action, ... Following the completion of the action, ...''
     
4) Preserve all evidence statements in meaning. You may rephrase only for fluency and redundancy reduction.

5) Output a single fluent paragraph.

\end{quote}

\section{Description Instructions}

\subsection{Instruction Diversification and Usage}
To reduce prompt sensitivity and improve robustness to instruction phrasing, we use an instruction diversification strategy during training. We prepare two instruction pools corresponding to the two clip types in PFED5+:
\begin{itemize}
    \item `Sit-at-rest' clips 
    \item Action clips 
\end{itemize}

For each clip type, we use ChatGPT-4o to create 20 semantically equivalent instruction variants, based on the base instruction and the prompt below.
\begin{quote}
Please generate 20 instructions that are semantically equivalent to \textit{[the  base instruction]} but vary in wording and structure. Ensure that each new instruction conveys the same core request but uses different phrasing and sentence construction.
\end{quote}

During training, one instruction is randomly sampled from the relevant bank for each clip at each iteration. At inference time, we always use the base instruction in the corresponding pool. Since the inference prompt is fixed, the instruction token embedding $E_t$ is constant during inference.

\subsection{`Sit-at-rest' Instructions}

\begin{enumerate}
    \item Provide a brief, objective summary of the person’s facial state and minor visible movements, focusing on the eyebrows, eyelids, cheeks, mouth corner, and mouth, and noting any additional details such as blinking, chin states, gaze shifts, or wrinkle changes. (\textbf{Base Instruction})
    \item Describe the person's facial appearance in a concise and neutral way, emphasizing the state of the eyebrows, eyelids, cheeks, mouth corners, and mouth, and include any subtle motions such as blinking, gaze direction, or wrinkle changes.
    \item Give a short, factual account of the person's facial expression and any slight movements, mentioning the eyebrows, eyelids, cheeks, corners of the mouth, and mouth, as well as other minor cues like blinking or shifts in gaze.
    \item Summarize briefly and objectively how the person's face appears, focusing on the position and tension of the eyebrows, eyelids, cheeks, and mouth region, and note if there are tiny movements such as blinks or eye direction changes.
    \item Provide an objective and succinct description of the person's facial condition, paying attention to the eyebrows, eyelids, cheeks, mouth corners, and mouth, and add observations on any minute actions like blinking or wrinkle variations.
    \item Write a concise summary of the face's current state, concentrating on the eyebrows, eyelids, cheeks, and mouth area, and point out any subtle visible movements such as a blink or a change in gaze.
    \item Offer a neutral, compact description of how the person's facial features appear, focusing on key areas-the eyebrows, eyelids, cheeks, and mouth-and mention minor activities like blinking or muscle twitches if visible.
    \item Describe, in a brief and impartial tone, the current state of the person's face, with attention to the eyebrows, eyelids, cheeks, corners of the mouth, and mouth, noting any small or brief movements.
    \item Give a short, objective outline of the facial state, including observations about the eyebrows, eyelids, cheeks, and mouth area, and mention any fine motions such as blinking or small shifts in gaze.
    \item Provide a compact and factual account of the facial expression, focusing on eyebrow shape, eyelid position, cheek form, and mouth posture, as well as any minimal visible movement like blinking or wrinkle formation.
    \item Write a brief, observation-based summary describing the person's facial appearance-covering the eyebrows, eyelids, cheeks, mouth corners, and mouth-and include minor dynamic details such as gaze movement or blinking.
    \item Present an objective and to-the-point description of the individual's face, noting the positions of the eyebrows, eyelids, cheeks, and mouth, and record any tiny motion such as eye blinks or chin shifts.
    \item Offer a concise report of the facial state, describing the eyebrows, eyelids, cheeks, and mouth area in a neutral tone, while also mentioning subtle physical indicators like blinking or wrinkle presence.
    \item Provide a short description that objectively summarizes the face's condition, including eyebrow placement, eyelid openness, cheek tension, and mouth shape, with attention to any faint movements.
    \item Summarize the person's facial appearance clearly and briefly, focusing on static features like eyebrows and cheeks, and any minor visible changes such as blinking or slight mouth movement.
    \item Describe succinctly the face's resting state, detailing the appearance of the eyebrows, eyelids, cheeks, and mouth, and mention any small visible actions like a blink or shift in gaze.
    \item Give a clear, unbiased summary of the person's current facial posture, emphasizing eyebrows, eyelids, cheeks, and the mouth area, and include minor cues such as blink frequency or wrinkle changes.
    \item Write a neutral, brief portrayal of the individual's face, concentrating on the eyebrows, eyelids, cheeks, and mouth region, with mention of tiny or barely noticeable movements.
    \item Produce a concise objective description of how the face looks, focusing on the eyebrows, eyelids, cheeks, and mouth corners, and noting any faint physical motion such as blinking or gaze drift.
    \item Provide a factual, succinct statement of the facial expression, covering the state of the eyebrows, eyelids, cheeks, and mouth, and mention any delicate or minimal visible movements.
\end{enumerate}

\subsection{Action Instructions}

\begin{enumerate}
  \item Provide an objective, phase-by-phase description of the face-pre-action, during-action, and post-action-highlighting the eyebrows, eyelids, cheeks, mouth corners, and lips, and mentioning any additional cues like blinking, chin movement, wrinkles, or gaze changes. (\textbf{Base Instruction})
  \item Describe the facial movement in up to three phases-before, during, and after the action-focusing on the eyebrows, eyelids, cheeks, mouth corners, and lips, and noting any other visible details such as blinking, chin position, wrinkles, or gaze shifts.
  \item Give a concise account of facial changes across up to three stages-before, during, and after the action-covering the eyebrows, eyelids, cheeks, mouth corners, and lips, and including any noticeable features such as blinking, chin motion, wrinkles, or gaze variation.
  \item Summarize the face across up to three stages-prior to, during, and following the action-describing the eyebrows, eyelids, cheeks, mouth corners, and lips, and noting other observable elements like blinking, chin shifts, wrinkle formation, or gaze movement.
  \item Offer an objective description of facial dynamics in up to three stages-before, during, and after the movement-focusing on eyebrows, eyelids, cheeks, mouth corners, and lips, and adding details like blinking, chin position, wrinkles, or gaze direction.
  \item Provide a stage-wise description of the facial state-pre-action, action, and post-action-covering eyebrows, eyelids, cheeks, mouth corners, and lips, and mentioning other visible traits such as blinking, chin adjustments, wrinkle changes, or gaze shifts.
  \item Describe the face across up to three sequential phases-before, during, and after the action-highlighting eyebrows, eyelids, cheeks, mouth corners, and lips, and noting other subtle indicators like blinking, chin motion, wrinkles, or eye gaze.
  \item Give an objective three-phase overview of the facial expression-before, during, and after the action-detailing eyebrows, eyelids, cheeks, mouth corners, and lips, along with additional observations like blinking, chin posture, wrinkles, or gaze changes.
  \item Offer a clear and factual summary of the facial movement in up to three parts-before, during, and after the action-mentioning eyebrows, eyelids, cheeks, mouth corners, and lips, and noting extra cues such as blinking, chin adjustments, wrinkles, or eye direction.
  \item Provide a structured account of facial behavior over up to three phases-before, during, and after the action-covering eyebrows, eyelids, cheeks, mouth corners, and lips, and adding any visible signs like blinking, chin movement, wrinkle development, or gaze variation.
  \item Describe facial activity in up to three temporal segments-before, during, and after the action-focusing on eyebrows, eyelids, cheeks, mouth corners, and lips, and including any other visible details like blinking, chin motion, wrinkles, or gaze shifts.
  \item Summarize facial changes in up to three phases-before, during, and after the action-addressing eyebrows, eyelids, cheeks, mouth corners, and lips, while also noting any other details such as blinking, chin position, wrinkles, or gaze orientation.
  \item Provide an objective breakdown of facial movement-before, during, and after the action-describing eyebrows, eyelids, cheeks, mouth corners, and lips, and mentioning any additional visible cues like blinking, chin posture, wrinkles, or gaze movement.
  \item Offer a concise, stage-based description of the facial action-pre-action, during-action, and post-action-covering eyebrows, eyelids, cheeks, mouth corners, and lips, and including other observable features such as blinking, chin shifts, wrinkles, or gaze changes.
  \item Give a stepwise description of the face through up to three stages-before, during, and after the action-mentioning eyebrows, eyelids, cheeks, mouth corners, and lips, and noting further details like blinking, chin position, wrinkle changes, or eye movement.
  \item Provide an objective depiction of facial changes across up to three phases-prior to, during, and following the action-focusing on eyebrows, eyelids, cheeks, mouth corners, and lips, and noting other perceptible elements like blinking, chin adjustments, wrinkles, or gaze shifts.
  \item Describe the facial expression over up to three stages-before, during, and after the action-highlighting eyebrows, eyelids, cheeks, mouth corners, and lips, and mentioning additional cues such as blinking, chin motion, wrinkles, or gaze direction.
  \item Offer a three-phase summary of facial movements-before, during, and after the action-covering eyebrows, eyelids, cheeks, mouth corners, and lips, and noting any other visible characteristics like blinking, chin shifts, wrinkles, or eye gaze changes.
  \item Give an objective overview of the facial sequence-before, during, and after the action-focusing on eyebrows, eyelids, cheeks, mouth corners, and lips, and adding extra observations such as blinking, chin movement, wrinkles, or gaze variation.
  \item Provide a factual description of the facial movement in up to three phases-before, during, and after the action-mentioning eyebrows, eyelids, cheeks, mouth corners, and lips, and including any other visible details such as blinking, chin adjustments, wrinkles, or gaze direction.
\end{enumerate}

\end{document}